\pdfoutput=1
\documentclass[11pt]{article}

\usepackage[]{emnlp2021}

\usepackage{times}
\usepackage{latexsym}

\usepackage[T1]{fontenc}
\usepackage[utf8]{inputenc}

\usepackage{microtype}

\usepackage{url}
\usepackage{tabularx}
\usepackage{booktabs}
\usepackage{graphicx}
\usepackage{subcaption}
\usepackage{makecell}
\title{Towards Realistic Single-Task Continuous Learning Research for NER}
\author{Justin Payan$^1$\thanks{Work completed while first author was an intern at Amazon Alexa AI.}, 
Yuval Merhav$^2$, 
He Xie$^2$, 
Satyapriya Krishna$^2$,\\
\textbf{Anil Ramakrishna$^2$, 
Mukund Sridhar$^2$, 
Rahul Gupta$^2$} \\
  $^1$University of Massachusetts Amherst, MA, USA \\
  $^2$Amazon Alexa AI, MA, USA \\
  $^1$\texttt{jpayan@umass.edu} \\
  $^2$\texttt{\{merhavy, hexie, satyapk, aniramak, harakere, gupra\}@amazon.com} \\
  }

\begin{document}
\maketitle
\begin{abstract}
  There is an increasing interest in continuous learning (CL), as
  data privacy is becoming a priority for real-world machine learning applications. Meanwhile, there
  is still a lack of academic NLP benchmarks that are applicable for realistic CL settings, which is a major challenge for the advancement of the field. In this paper we discuss some of the unrealistic data characteristics of public datasets, study the challenges of realistic single-task continuous learning as well as the effectiveness of data rehearsal as a way to mitigate accuracy loss. We construct a CL NER dataset from an existing publicly available dataset and release it along with the code to the research community\footnote{\url{https://github.com/justinpayan/StackOverflowNER-NS}}.  
\end{abstract}

\section{Introduction}

Data privacy is a hot topic in ML, gaining attention in both industry and academia~\cite{papernot2016towards, perera2015big}. One of the topics of interest is data retention, which can be improved by training models incrementally~\cite{wu2019large}. An ideal training regime would involve continuously updating a model on newly acquired data, then deleting the data. Benchmarking CL strategies today is still highly nonstandard in academic research~\cite{maltoni2019continuous}. 

One key difference between real-world and academic datasets is the dynamic nature of the former. Academic datasets are often static and contain data that is annotated all at once based on fixed annotation guidelines. When building real-world applications, such data collection and annotation workflow is often not realistic. Rather, an initial dataset is created and then is evolved over time based on usage pattern changes and business needs. For example, new labels are added periodically, data distribution changes significantly due to seasonality or other factors,  annotation guidelines are updated, etc. While such datasets exist in industry, they are often confidential or proprietary and cannot be shared with the research community. 

Consequently, the academic CL research focus has been mainly on the multi-task learning scenario, where the same model is required to learn a number of isolated tasks incrementally without forgetting how to solve the previous ones. In this work we tackle the single-task scenario using the Named Entity Recognition (NER) task. There is only one task, but it evolves over time due to data distribution shift, introduction of new labels, or other factors. Single-task is often considered to be more difficult than multi-task~\cite{kemker2018measuring,kemker2018fearnet,maltoni2019continuous} and is also a common real-world scenario. 

To the best of our knowledge, there are no public NLP benchmarks specifically designed for single-task CL. In order to study this problem we pick the recent StackOverflowNER dataset~\cite{tabassum-etal-2020-code}. The dataset authors' motivation was studying named entity recognition in the social computer programming domain, not continuous learning. However, the characteristics of the dataset are ideal for a study in CL. It spans roughly 10 years (from September 2008 to March 2018) of question-answer threads that are manually annotated with close to 30 types of entities. The dataset is also very diverse and has a large sample size -- other public NER datasets are too small or contain only a few entity types. Finally, the manual annotation process resembles that of industrial use cases, where the labeling process might be subject to noise and human error.

\begin{figure*}[!ht]
     \centering
     \begin{subfigure}[!t]{1.0\textwidth}
         \centering
         \includegraphics[width=\textwidth]{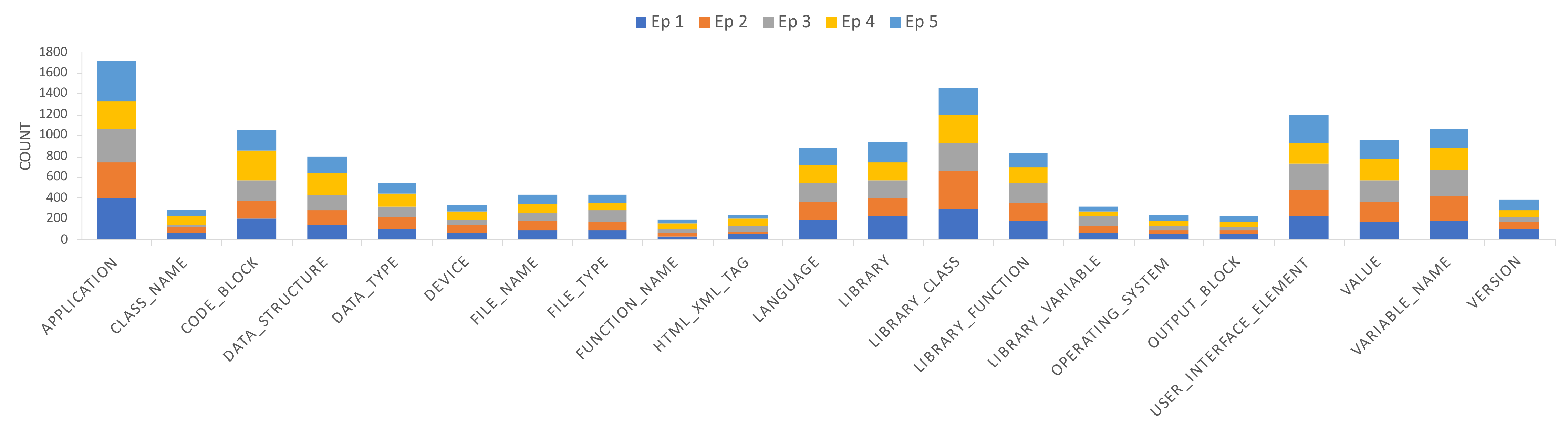}
         \caption{Temporal}
         \label{fig:orig-dist}
     \end{subfigure}
     \hfill
     \begin{subfigure}[!t]{1.0\textwidth}
         \centering
         \includegraphics[width=\textwidth]{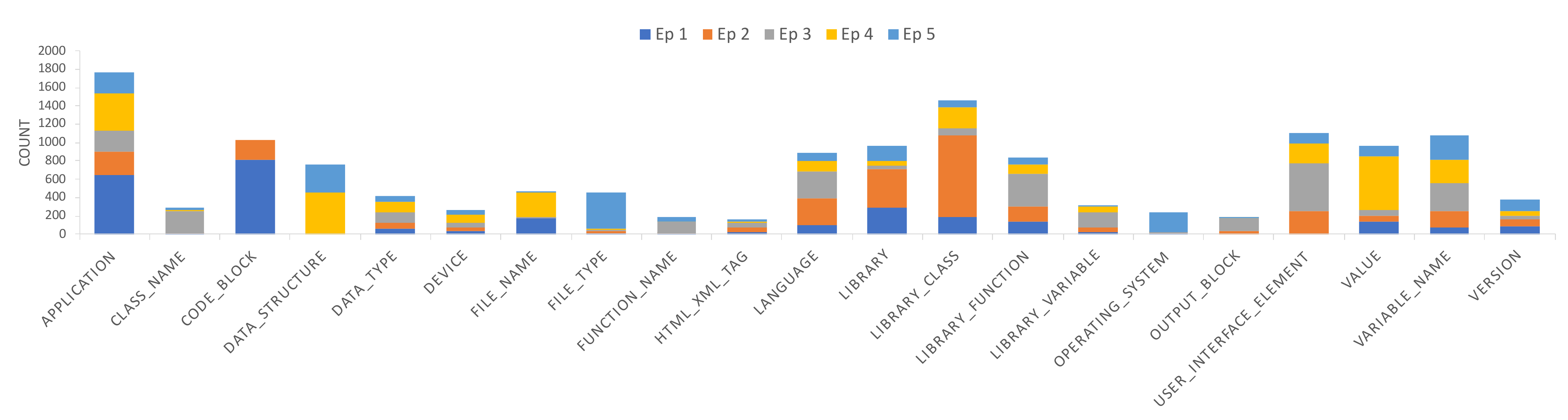}
         \caption{Skewed}
         \label{fig:skewed-dist}
     \end{subfigure}
        \caption{Entity type distribution across episodes comparing the temporal and skewed episodes. Each vertical bar has the frequency for each of the 5 episodes. For readability we removed types with low counts in each episode.}
        \label{fig:three graphs}
\end{figure*}

In order to simulate CL we split the data into time-based episodes and train an NER model incrementally over 5 episodes.  Our results show no regression and limited forgetting. To present a more realistic challenge, we propose a configurable distribution-based sampling of data inspired by our experiences with a confidential industrial dataset. We show that our sampled episodes can be used to study the effectiveness of different single-task CL strategies in the context of NER. The resulting dataset is the main contribution of this work.

\section{Continuous Learning}

\textbf{Strategies}. The main focus in training deep learning models in CL fashion is prevention of catastrophic forgetting~\cite{kirkpatrick2017overcoming}. Neural networks trained on new data tend to do poorly on old data and to mitigate catastrophic forgetting different strategies have been proposed, such as specific architectures for CL~\cite{pmlr-v78-lomonaco17a,rusu2016progressive}, regularization techniques~\cite{kirkpatrick2017overcoming,li2017learning}, and data rehearsal/replay where small subsets of old data (real or generated) is periodically supplied to the model during training on new data~\cite{sun2019lamol,shin2017continual}. The latter is considered a strong CL baseline~\cite{maltoni2019continuous} and thus we use this approach in this study. We also compare against a variation of the replay-based GDumb baseline~\cite{prabhu2020gdumb}. GDumb collects examples into a memory buffer with a limited budget size $k$, balancing the distribution over labels by greedily sampling under-represented label types and ejecting over-sampled label types. The model trains on the buffer after all tasks are seen.

\textbf{Our CL model}. Our model design is inspired by LAMOL~\cite{sun2019lamol} and adapted for NER.  We employ a pre-trained GPT-2 language model base~\cite{radford2019language}, then 2 layers of bi-LSTM with 768 dimensions in each direction, a $\tanh$ non-linearity and linear transformation (1536 by number of labels), and a CRF layer to predict labels. All parameters besides the GPT-2 base (pre-trained on OpenAI's WebText) are randomly initialized, and we train or finetune all parameters during training. Training on all 5 episodes takes less than 12 hours on an NVIDIA Tesla M40 GPU for all experimental settings. We assume that all entity types are known in advance so we do not need to expand the label size in a later episode if a new label is introduced. In our experiments, our baseline is a model fine-tuned on all training data. We compare the baseline to GDumb and two CL strategies: training with and without data replay.

\textbf{Data replay}.  For each episode (barring the first), we set the size of replayed examples to be sampled from previous episodes to 20\% of the size of the current episode’s training set. An equal number of replayed examples are sampled from each previous episode. To apply GDumb to NER, we add examples containing under-represented entity types to the buffer, and we eject examples which have the maximum value for their least well-represented entity type. 

\section{Experimental Setup}

\subsection{Time-Based Episodic Setup}

Our first motivation is to investigate continual learning over time.  We construct our continual learning datasets from StackOverflowNER, a dataset of questions and answers on StackOverflow annotated with 28 entity types~\cite{tabassum-etal-2020-code}. We combine StackOverflowNER's training and development sets to construct a pool for sampling training episodes, and we use the test set as a pool for sampling test episodes. All data splits and code are available at~\url{https://github.com/justinpayan/StackOverflowNER-NS}.

\begin{table}[!hp]
    \centering
        \begin{tabularx}{\columnwidth}{ccc}
        \toprule
       \small Episode & \small Date Range & \small Train / Test Size\\
       \midrule
        1& 8/4/2008 -- 6/26/2012 & 2551 / 775 \\
         2& 6/27/2012 -- 3/13/2014 & 2444 / 665 \\
          3& 3/14/2014 -- 6/27/2015 & 2243 / 521 \\
           4& 6/28/2015 -- 10/1/2016 & 2450 / 496 \\
            5& 10/2/2016 -- 3/27/2018 & 2386 / 632 \\
        %\midrule

        \bottomrule
        \end{tabularx}
        \caption{Date boundaries for each episode.}
    \label{tab:dataset_date_ranges}
    \end{table}

We first split the StackOverflowNER data into 5 time-based episodes. The StackOverflowNER dataset does not have timestamps, so we align their annotated examples with posts in the StackOverflow data dump. We select date boundaries for each episode to obtain roughly equal-sized training and test sets. Table \ref{tab:dataset_date_ranges} lists the dates. 

\subsection{Results}

Figure \ref{fig:orig-dist} shows the distribution of each entity type across the 5 episodes. While some entity types are more common than others, the frequency distribution is consistent across episodes. The percentage of examples tagged with a particular entity type does not change much across episodes and there are no deletions or additions of new entity types over time. Such data characteristics are not realistic for a real-world application evolving over 10 years.

We train our model incrementally on the 5 episodes with and without data replay and compare it to a baseline model that is trained on all data at once in a non-CL fashion. Table \ref{tab:so-all-results} shows the averaged F1 score over the 5 episodes' test data (comprehensive results can be found in Appendix~\ref{sec:comp_results}). Not surprisingly, training incrementally performs on-par with training on all data at once, meaning that if there is any catastrophic forgetting, it does not impact the test performance of the model. As such, applying data replay that is supposed to mitigate catastrophic forgetting has no benefit and even results in a mild performance degradation. Preliminary manual analysis suggests that degredation stems from memorization of infrequent patterns sampled in the relatively small replay set. Given these data characteristics and results, it is clear that the dataset, in this format, is not proper for a comparison of CL strategies.

\subsection{Skewed Class Distribution Setup}

\begin{table*}[]
    \centering
        \begin{tabular}{*{5}{c}}
        \toprule
        & &  Overall & CodeBlock &  DataStruct. \\
        \midrule
        & Baseline (non-CL) & 51.36 & 25.67 & 75.27 \\
        Temporal & CL w/o Replay & 51.52 & 28.59 & 73.76 \\
        & CL w/ Real Replay & 51.12 & 26.41 & 72.82 \\
        \midrule
        &  Baseline (non-CL) & 52.24 & 12.51  & 32.03 \\
        &  CL w/o Replay & 42.61 & 0.00 & 32.60 \\
         Skewed & CL w/ Replay  & 49.82 & 7.74 & 33.82 \\
        &  GDumb (500) & $24.28 \pm 0.98$ & $6.81 \pm 0.49$ & $7.80 \pm 4.25$ \\
        &  GDumb (1000) & $35.41 \pm 0.90$ & $8.10 \pm 0.60$ & $24.09 \pm 1.38$ \\
        &  GDumb (1500) & $40.19 \pm 0.67$ & $8.82 \pm 0.54$ & $27.46 \pm 1.52$ \\
        \bottomrule
        \end{tabular}
        \caption{Overall and selected entity type F1 scores after training incrementally over all 5 episodes vs on all training data at once. All scores are averaged over all 5 episodes' test sets. We also compare against the GDumb baseline, with memory budgets of 500, 1000, or 1500 examples. We run GDumb over 10 random orderings within each episode, and report means and standard deviations.}
    \label{tab:so-all-results}
\end{table*} 

Motivated by our findings, we create an updated version of the episodic dataset based on more realistic assumptions. The first assumption is of data distribution shift and variance. Data distribution shift is expected due to various factors such as seasonality. A second factor is annotation cost. When a model is doing well on specific types of data/labels, there is no need to continue annotating similar examples and labels.
% This can be a problem if a model is trained incrementally and suffers from catastrophic forgetting. 
We modify the StackOverflow dataset by sampling the distribution over entity types from a Dirichlet distribution for each episode. To simplify, we assume independence between entity types, although entity types often co-occur.

We first compute the distribution over entity types in the training pool, and denote that with $\mathbf{\alpha}$. We then sample distributions for the 5 training episodes, $\{\mathbf{X}_i^{tr}\}_{i=1}^5 \sim Dir(c\mathbf{\alpha})$ and the 5 test episodes $\{\mathbf{X}_i^{te}\}_{i=1}^5 \sim Dir(\mathbf{X}_i^{tr})$. We set $c=5$ but the parameter can be changed to increase or decrease variance. To sample the train (test) episodes, we cycle through the episodes, each time selecting an entity type from the episode's distribution and then selecting an example containing that entity type from the train (test) pool without replacement.

In addition to modeling distribution shift, we also introduce class incrementality. We select 3 entity types that are relatively frequent: \textsc{Code\_Block}, \textsc{Data\_Structure}, and \textsc{User\_Interface\_Element}. We simulate the data shift by removing the \textsc{Code\_Block} entity in episode 3 and onward, adding the \textsc{Data\_Structure} entity only in episodes 4 and 5, and removing the \textsc{User\_Interface\_Element} entity from episode 1. To achieve this, each time we sample one of these entity types in a disallowed episode, we put that sample back into the pool.

\subsection{Results}

Figure \ref{fig:skewed-dist} shows the distribution of each entity type across the 5 skewed episodes. In comparison to Figure \ref{fig:orig-dist}, one can see the increased variance of the distribution across episodes. Appendix~\ref{sec:diachronicity} shows further comparisons between the skewed and temporal settings. We find the degree of variance to be similar to that of our confidential industrial NER dataset.
% For example, \textsc{Library\_Class} appears 895 times in episode 2 but only 80 times in the next episode.  
Following the previous model training procedure, we train our model incrementally on the 5 skewed episodes with and without data replay and compare it to a baseline model that is trained on all data at once in a non-CL fashion. Table \ref{tab:so-all-results} shows the averaged F1 score over the 5 episodes' test data. Contrary to the previous setup, we see that the non-CL baseline heavily outperforms CL without replay. Data replay helps, but there is still a gap in performance. Even with a buffer size of 1500, GDumb greatly underperforms even the continual learning setup without replay. As GDumb is a strong baseline, this suggests the setting is quite difficult.

We can also see the impact of excluding \textsc{Code\_Block} from episode 3 onward. The model completely stops predicting it in the no replay case. The CL models also struggle with \textsc{Data\_Structure}, perhaps because the final model learns a grossly inflated probability for that tag while the baseline sees the training examples in a consistently balanced fashion.

We find that the CL models suffer from subtler distribution shift errors too. For example, we see forgetting of common named entities. Episode 1 includes many instances with the \textsc{Application} ``Android Studio,'' while Episode 5 only references the \textsc{Operating\_System} ``Android.'' Thus the final CL models classify ``Android'' as \textsc{Operating\_System} and ``Studio'' as \textsc{Application}. More sophisticated replay techniques could address such issues by reducing distribution shift or replaying representatives for common entities/phrases.
% A training example from Episode 1 is ``[a]s this needs to be done for each date, it can be done using cursor,'' with ``cursor'' labeled as \textsc{Code\_Block}. Both CL models classify ``cursor'' as \textsc{Library\_Function} instead of \textsc{Code\_Block}, presumably because \textsc{Code\_Block} is not seen in Episodes 3, 4, and 5. Even with replay, the model mistakenly learns that other types are more common than \textsc{Code\_Block}. 

\subsection{Forgetting Over Time}
\begin{figure*}[!ht]
    \centering

    \begin{subfigure}[t]{0.48\textwidth}
        \centering
        \includegraphics[width=\textwidth]{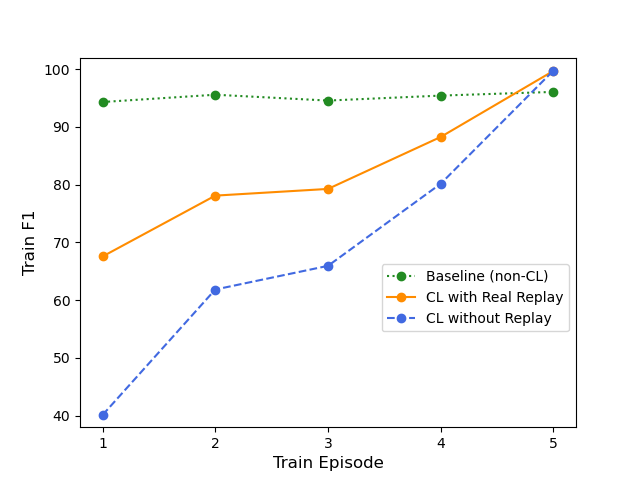}
        \caption{Skewed Train}
        \label{fig:forget-on-train}
    \end{subfigure}
    \hfill
    \begin{subfigure}[t]{0.48\textwidth}
        \centering
        \includegraphics[width=\textwidth]{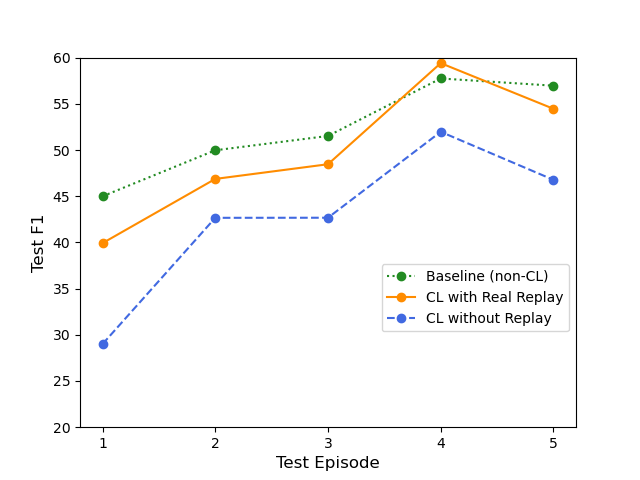}
        \caption{Skewed Test}
        \label{fig:forget-on-test}
    \end{subfigure}

    \begin{subfigure}[t]{0.48\textwidth}
        \centering
        \includegraphics[width=\textwidth]{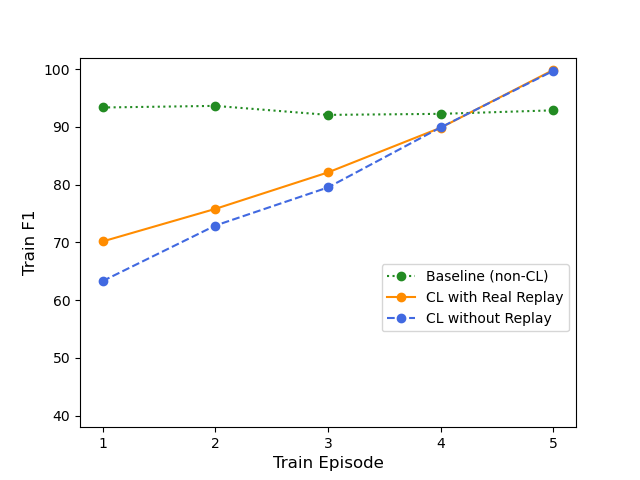}
        \caption{Temporal Train}
        \label{fig:forget-on-train-temp}
    \end{subfigure}
    \hfill
    \begin{subfigure}[t]{0.48\textwidth}
        \centering
        \includegraphics[width=\textwidth]{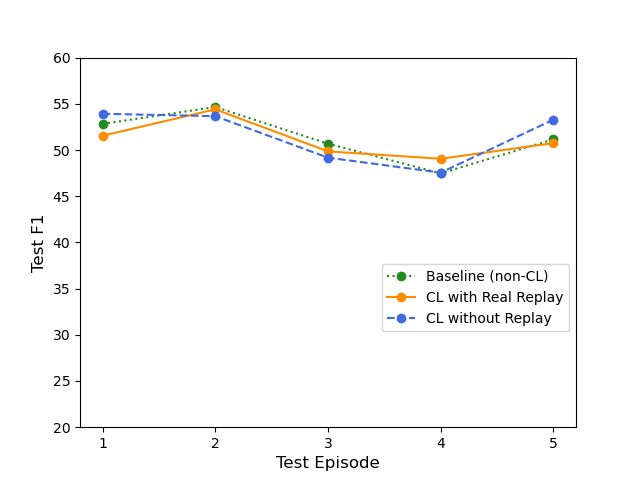}
        \caption{Temporal Test}
        \label{fig:forget-on-test-temp}
    \end{subfigure}

       \caption{Overall F1 score evaluated on each of the 5 episodes' train or test sets, for both skewed and temporal settings. All models evaluated here are trained on data from all episodes, where the CL models are trained incrementally, starting with episode 1 and finishing with episode 5.}
       \label{fig:forgetting_plots}
\end{figure*}

% \begin{figure}
%     \centering
%     \begin{subfigure}[t]{0.45\textwidth}
%         \centering
%         \includegraphics[width=\linewidth]{example-image-a.pdf} 
%         \caption{Generic} \label{fig:timing1}
%     \end{subfigure}
%     \hfill
%     \begin{subfigure}[t]{0.45\textwidth}
%         \centering
%         \includegraphics[width=\linewidth]{example-image-b.pdf} 
%         \caption{Competitors} \label{fig:timing2}
%     \end{subfigure}

%     \vspace{1cm}
%     \begin{subfigure}[t]{\textwidth}
%     \centering
%         \includegraphics[width=\linewidth]{example-image-c.pdf} 
%         \caption{Price regulation} \label{fig:timing3}
%     \end{subfigure}
%     \caption{Some general caption of all the figures. In (\subref{fig:timing1}) you can see a  green square....}
% \end{figure}

Figure~\ref{fig:forget-on-train} shows how the \textbf{final} model (trained on all data) in each experiment performs on each of the train episodes with the skewed distribution. The figure shows that the CL approaches suffer from catastrophic forgetting compared to the non-CL baseline, with no replay performing worse, as expected. While the performance of the baseline model is consistent over the train episodes, the CL models' performance degrades on the earlier training episodes. While data replay helps, the gap is still large which leaves room for future work. The same plot for the temporal data splits is shown in Figure~\ref{fig:forget-on-train-temp}. Forgetting still occurs in this case, but at a lower rate. 

We also demonstrate the forgetting on the test sets in Figures~\ref{fig:forget-on-test} and~\ref{fig:forget-on-test-temp}, where we see little impact of forgetting for the temporal setting compared to the skewed setting. The baseline's lower performance on skewed episodes 1, 2, and 3 stems from the removal of \textsc{User\_Interface\_Element} from test episode 1 and \textsc{Data\_Structure} from test episodes 1, 2, and 3. The baseline can predict these entity types with relatively high accuracy, and they are fairly common. When they are removed, the baseline model loses the boost in overall F1 these types provide. Overall, we see higher forgetting when evaluating the CL approaches on train than on test, which can be explained by overfitting to the most recent episodes during training.

In the future we would like to explore hyper-parameter tuning which could further reduce forgetting, and apply privacy preserving techniques such as generative replay~\cite{sun2019lamol}. Establishing more advanced benchmarks using recent CL techniques or creating similar episodic splits for other NLP tasks would also be of interest.

\section{Conclusions}
We demonstrate that even in an academic dataset spanning a decade, some important characteristics of applied single-task continual learning settings, such as data shift and label imbalance, are missing. We modify and release a dataset that contains some of these realistic challenges, and we establish a data replay baseline. Although the ability to access and publish statistics for real industrial datasets is limited due to privacy and business concerns, we find that our dataset exhibits many important similarities to such datasets. Our method for producing the dataset is configurable and can be used to build different degrees of data variance to support different use cases. Although our dataset is a useful first step towards more realistic single-task continual learning, this work highlights the need for a public benchmark with truly continuous annotation. 

\section*{Acknowledgements}

We are grateful to Emre Barut for helpful feedback on drafts of this paper.

\bibliography{emnlp2021}
\bibliographystyle{acl_natbib}

\newpage
\clearpage

\appendix

\section{Comprehensive Results}
\label{sec:comp_results}

We include full results for all entity types, for both the temporal data split and the skewed data split. The full results for the temporal data split are included in Table~\ref{tab:comp_temp}, and the full results for the skewed data split are included in Table~\ref{tab:comp_skew}.

\begin{table*}[!tp]
\centering
    \begin{tabular}{*{5}{c}}
    \toprule
   Entity Type & \begin{tabular}{c}Baseline \\(non-CL)\end{tabular} & \begin{tabular}{c}CL w/o \\Replay\end{tabular} & \begin{tabular}{c}CL w/ \\Real Replay\end{tabular} & Avg. Count \\

    \midrule
    \small Overall & 51.36 & 51.52 & 51.12 & 777.60 \\
    \small Algorithm & 24.00 & 19.64 & 21.82 & 0.00 \\
    \small Application & 57.94 & 57.76 & 58.36 & 2.80 \\
    \small ClassName & 25.38 & 18.89 & 18.33 & 80.00 \\
    \small CodeBlock & 25.67 & 28.59 & 26.41 & 25.80 \\
    \small DataStructure & 75.27 & 73.76 & 72.82 & 59.80 \\
    \small DataType & 67.52 & 70.24 & 70.81 & 48.00 \\
    \small Device & 59.38 & 60.24 & 58.99 & 21.60 \\
    \small ErrorName & 3.64 & 3.64 & 14.16 & 10.60 \\
    \small FileName & 62.31 & 64.06 & 60.01 & 3.60 \\
    \small FileType & 69.82 & 66.28 & 77.19 & 32.60 \\
    \small FunctionName & 12.25 & 4.86 & 9.71 & 21.60 \\
    \small HTMLXMLTag & 42.32 & 41.51 & 40.96 & 9.20 \\
    \small KeyboardIP & 1.74 & 9.78 & 1.67 & 10.40 \\
    \small Language & 75.41 & 74.09 & 70.75 & 7.00 \\
    \small Library & 53.97 & 53.28 & 47.46 & 35.40 \\
    \small LibraryClass & 47.55 & 48.00 & 47.06 & 50.20 \\
    \small LibraryFunction & 44.69 & 48.43 & 47.20 & 72.80 \\
    \small LibraryVariable & 18.58 & 10.79 & 21.57 & 43.00 \\
    \small License & 0.00 & 0.00 & 0.00 & 21.80 \\
    \small OperatingSystem & 82.46 & 79.15 & 82.85 & 0.00 \\
    \small Organization & 10.00 & 53.33 & 43.33 & 12.20 \\
    \small OutputBlock & 75.20 & 68.71 & 67.14 & 1.80 \\
    \small UserInterfaceElement & 56.43 & 56.96 & 56.67 & 10.80 \\
    \small UserName & 35.83 & 35.69 & 32.21 & 69.40 \\
    \small Value & 45.68 & 44.88 & 34.68 & 4.60 \\
    \small VariableName & 28.44 & 28.81 & 27.53 & 43.00 \\
    \small Version & 72.05 & 72.26 & 72.74 & 53.00 \\
    \small Website & 25.99 & 22.29 & 28.00 & 21.20 \\
    
    \bottomrule
    \end{tabular}
    \caption{F1 scores by type after training incrementally over all 5 temporal episodes vs on all training data at once. Scores are averaged over all 5 episodes' test sets. We also denote the average count of each entity type in all 5 test episodes.}
\label{tab:comp_temp}
\end{table*}
\begin{table*}[!tp]
    \centering
        \begin{tabular}{*{6}{c}}
        \toprule
    %    Entity Type & Baseline (non-CL) & CL w/o Replay & CL w/ Real Replay & GDumb (1500) & Avg. Count \\
       Entity Type & \begin{tabular}{c}Baseline \\(non-CL)\end{tabular} & \begin{tabular}{c}CL w/o \\Replay\end{tabular} & \begin{tabular}{c}CL w/ \\Real Replay\end{tabular} & \begin{tabular}{c}GDumb \\(1500)\end{tabular} & Avg. Count \\
        \midrule
        \small Overall & 52.24 & 42.61 & 49.82 & $40.19 \pm 0.67$ & 750.40 \\
\small Algorithm & 10.00 & 14.44 & 28.33 & $32.43 \pm 4.17$ & 0.00 \\
\small Application & 55.93 & 53.01 & 55.68 & $47.23 \pm 1.33$ & 3.20 \\
\small ClassName & 19.84 & 5.24 & 8.21 & $21.17 \pm 1.94$ & 75.40 \\
\small CodeBlock & 12.51 & 0.00 & 7.74 & $8.82 \pm 0.54$ & 25.60 \\
\small DataStructure & 32.03 & 32.60 & 33.82 & $27.46 \pm 1.52$ & 47.20 \\
\small DataType & 72.45 & 67.24 & 68.77 & $63.53 \pm 3.89$ & 45.60 \\
\small Device & 53.32 & 47.39 & 47.21 & $45.28 \pm 5.18$ & 21.20 \\
\small ErrorName & 0.00 & 0.00 & 10.00 & $4.39 \pm 2.61$ & 10.60 \\
\small FileName & 54.79 & 6.17 & 46.81 & $39.09 \pm 3.88$ & 3.60 \\
\small FileType & 55.95 & 38.51 & 55.10 & $43.37 \pm 5.23$ & 32.60 \\
\small FunctionName & 26.16 & 5.34 & 8.58 & $11.23 \pm 2.03$ & 25.80 \\
\small HTMLXMLTag & 40.25 & 27.98 & 41.61 & $33.55 \pm 3.70$ & 9.20 \\
\small KeyboardIP & 8.00 & 5.71 & 13.33 & $8.31 \pm 3.39$ & 10.40 \\
\small Language & 69.11 & 67.59 & 67.83 & $57.26 \pm 1.21$ & 7.00 \\
\small Library & 55.35 & 48.26 & 55.43 & $40.70 \pm 2.57$ & 35.60 \\
\small LibraryClass & 48.52 & 42.13 & 45.70 & $36.97 \pm 2.20$ & 51.40 \\
\small LibraryFunction & 44.95 & 34.87 & 44.49 & $32.91 \pm 3.64$ & 75.40 \\
\small LibraryVariable & 23.33 & 7.99 & 3.12 & $6.21 \pm 1.99$ & 41.40 \\
\small License & 0.00 & 0.00 & 0.00 & $0.00 \pm 0.00$ & 22.40 \\
\small OperatingSystem & 79.74 & 60.45 & 65.81 & $64.81 \pm 3.93$ & 0.00 \\
\small Organization & 13.33 & 20.00 & 20.00 & $21.76 \pm 4.62$ & 13.20 \\
\small OutputBlock & 63.07 & 0.00 & 63.78 & $59.16 \pm 6.16$ & 2.00 \\
\small UserInterfaceElement & 44.94 & 40.81 & 43.25 & $34.68 \pm 2.02$ & 10.60 \\
\small UserName & 30.73 & 39.63 & 36.00 & $28.48 \pm 4.81$ & 54.00 \\
\small Value & 56.27 & 45.75 & 46.19 & $38.92 \pm 2.61$ & 4.60 \\
\small VariableName & 25.87 & 26.05 & 26.35 & $18.83 \pm 3.00$ & 42.80 \\
\small Version & 77.89 & 70.54 & 77.41 & $73.98 \pm 3.18$ & 51.60 \\
\small Website & 36.66 & 27.81 & 49.53 & $35.05 \pm 4.63$ & 22.20 \\
        \bottomrule
        \end{tabular}
        \caption{F1 scores by type after training incrementally over all 5 skewed episodes vs on all training data at once. Scores are averaged over all 5 episodes' test sets. We also include results for GDumb with memory budget of 1500 examples, averaged over 10 random initializations. We also denote the average count of each entity type in all 5 test episodes.}
    \label{tab:comp_skew}
    \end{table*}

\section{Diachronicity of Temporal and Skewed}
\label{sec:diachronicity}

\begin{table*}[]
    \centering
        \begin{tabular}{*{6}{c}}
        \toprule
        & \small Ep. 1 & \small Ep. 2 & \small Ep. 3 & \small Ep. 4 & \small Ep. 5 \\
        \midrule
        & \small Application & \small LibraryClass & \small Application & \small CodeBlock & \small Application\\
        \small Temporal & \small LibraryClass & \small Application & \small LibraryClass & \small LibraryClass & \small UserInterfaceElem.\\
        \small Train & \small UserInterfaceElem. & \small UserInterfaceElem. & \small UserInterfaceElem. & \small Application & \small LibraryClass\\
         & \small Library & \small VariableName & \small VariableName & \small VariableName & \small Library\\
         & \small CodeBlock & \small Value & \small Value & \small Value & \small CodeBlock\\
        \midrule
         & \small UserInterfaceElem. & \small UserInterfaceElem. & \small LibraryClass & \small LibraryClass & \small Application\\
        \small Temporal & \small Application & \small LibraryClass & \small Value & \small Application & \small CodeBlock\\
        \small Test & \small LibraryClass & \small Application & \small CodeBlock & \small Library & \small VariableName\\
         & \small VariableName & \small LibraryFunction & \small VariableName & \small CodeBlock & \small LibraryFunction\\
         & \small Library & \small LibraryVariable & \small DataStructure & \small UserInterfaceElem. & \small Library\\
        \midrule
         & \small CodeBlock & \small LibraryClass & \small UserInterfaceElem. & \small Value & \small FileType\\
         \small Skewed & \small Application & \small Library & \small LibraryFunction & \small DataStructure & \small DataStructure\\
         \small Train & \small Library & \small Language & \small Language & \small Application & \small VariableName\\
         & \small LibraryClass & \small Application & \small VariableName & \small FileName & \small Application\\
         & \small FileName & \small UserInterfaceElem. & \small ClassName & \small VariableName & \small OperatingSystem\\
        \midrule
         & \small CodeBlock & \small UserInterfaceElem. & \small VariableName & \small DataStructure & \small LibraryClass\\
         \small Skewed & \small Value & \small Language & \small UserInterfaceElem. & \small Application & \small DataStructure\\
         \small  Test & \small Application & \small CodeBlock & \small Application & \small LibraryClass & \small VariableName\\
         & \small Library & \small Application & \small ClassName & \small LibraryFunction & \small Library\\
         & \small LibraryClass & \small FileType & \small LibraryClass & \small UserInterfaceElem. & \small FileName\\
        
            \bottomrule
        \end{tabular}
        \caption{Top five entity types (in order) for each episode of temporal/skewed train/test splits.}
    \label{tab:top5}
\end{table*}
\begin{table*}[!tp]
    \centering
        \begin{tabular}{*{2}{c}}
        \toprule
         & \begin{tabular}{c} Instead, start a \textbf{command prompt (Application)}  and \\ " \textbf{cd (Code\_Block)} " to where your \textbf{jar (File\_Type)} file is. \end{tabular}  \\\\
        \textsc{Code\_Block} & \begin{tabular}{c} Add \textbf{rm -r (Code\_Block)} to remove the \\ file hierarchy rooted in each file argument. \end{tabular} \\\\
        &  \textbf{rm /path/to/directory/ * (Code\_Block)} \\
        \midrule
        & \begin{tabular}{c} Allocate an \textbf{array (Data\_Structure)} of \\ \textbf{pointers (Data\_Type)} to \textbf{chars (Data\_Type)} \end{tabular} \\\\
        \textsc{Data\_Structure} & \begin{tabular}{c} where \textbf{keywords (Variable\_Name)} is the \\ \textbf{list (Data\_Structure)} of \textbf{strings (Data\_Type)} \\
            so we can parse and find the correct item, \\ and \textbf{session (Variable\_Name)} is the \\ a new \textbf{session (Library\_Class)} \\ from the \textbf{requests (Library)} module. \end{tabular}  \\\\
        & \begin{tabular}{c} I need to get the 14 days average \\ \textbf{Col 1 (Variable\_Name)} and update \textbf{Col 2 (Variable\_Name)} \\ of the same \textbf{table (Data\_Structure)}. \end{tabular}\\
        \midrule
        &  \begin{tabular}{c} There will be a class method, \\ which opens a new \textbf{tab (User\_Interface\_Element)}, \\ renders some \textbf{HTML (Language)}, \\ and returns the \textbf{PDF (File\_Type)} data, \\ and closes the \textbf{tab (User\_Interface\_Element)}. \end{tabular} \\\\
        \textsc{User\_Interface\_Element} & \begin{tabular}{c} I'm trying to create a responsive effect, \\ where I hide a \textbf{column (User\_Interface\_Element)} \\ when my \textbf{screen (User\_Interface\_Element)} \\ is \textbf{960 (Value)} or lower.  \end{tabular} \\\\
        & \begin{tabular}{c} But in \textbf{iOS (Operating\_System)} \textbf{10 (Version)}, \\ \textbf{photos (User\_Interface\_Element)} not appearing until I tap on \\ \textbf{cell (User\_Interface\_Element)} that holds \\ \textbf{collection view (Library\_Class)}. \end{tabular} \\
        \bottomrule
        \end{tabular}
        \caption{Examples containing the \textsc{Code\_Block}, \textsc{Data\_Structure}, and \textsc{User\_Interface\_Element} types. We remove all examples with these types in different episodes to simulate class incrementality in the skewed dataset. All entities are bolded with the entity type in parentheses following the entity.}
    \label{tab:diachron_exs}
    \end{table*}

We include some additional demonstrations of the differences between the temporal and skewed settings. In Table~\ref{tab:top5}, we show the top five entity types for all episodes' train and test for both settings. Although there is some variation across episodes for the temporal setting, the variation is stronger for the skewed setting.

We demonstrate a few examples of the \textsc{Code\_Block}, \textsc{Data\_Structure}, and \textsc{User\_Interface\_Element} types in Table~\ref{tab:diachron_exs}. Recall that in the skewed data, we remove the \textsc{Code\_Block} entity in episode 3 and onward, add the \textsc{Data\_Structure} entity only in episodes 4 and 5, and remove the \textsc{User\_Interface\_Element} entity from episode 1. This behavior impacts the top five entities, as Table~\ref{tab:top5} makes apparent.

\end{document}